\pdfoutput=1

\documentclass[11pt]{article}

\usepackage[final]{acl}
\usepackage{times}
\usepackage{latexsym}

\usepackage[T1]{fontenc}

\usepackage[utf8]{inputenc}

\usepackage{microtype}
\usepackage[compact]{titlesec}

\usepackage{inconsolata}

\usepackage{graphicx}

\usepackage{xcolor}

\usepackage{placeins}

\usepackage{float} 

\usepackage{amsmath}

\usepackage{pdfpages} 

\usepackage{graphicx}  

\usepackage{booktabs}

\usepackage{multirow}

\usepackage{enumitem}

\definecolor{text_tokens}{RGB}{0, 200, 168}
\definecolor{prompt_pads}{RGB}{227, 133, 45}
\definecolor{blank_pads}{RGB}{230, 178, 126}

\definecolor{blue_fig_4}{RGB}{85, 111, 175}
\definecolor{orange_fig_4}{RGB}{203, 133, 95}
\definecolor{green_fig_4}{RGB}{84, 173, 87}
\definecolor{red_fig_4}{RGB}{180, 77, 82}

\newcommand{\pad}{\texttt{pad}}

\newcommand{\encoderMethod}{ITE}
\newcommand{\diffusionMethod}{IDP}

\title{Padding Tone - Revealing the role of the padding tokens in T2I Models}

\title{Decoding the Padding Tone: Revealing Padding Tokens Role in T2I Models}

\title{Padding Tone: A Mechanistic Analysis of Padding Tokens in T2I Models}

\author{
Michael Toker$^1$ \hspace{1em} Ido Galil$^{1,2}$ \hspace{1em} Hadas Orgad$^1$ \hspace{1em} {\bf Rinon Gal}$^2$ \hspace{1em} {\bf Yoad Tewel}$^{2,4}$ \\
{\bf Gal Chechik}$^{2,3}$ \hspace{1em} {\bf Yonatan Belinkov}$^1$ \\
$^1$Technion -- Israel Institute of Technology \hspace{2em} $^2$NVIDIA \\ \hspace{2em} $^3$Bar-Ilan University \hspace{1em} $^4$Tel-Aviv University \\
        \texttt{\{tok,orgad.hadas\}@campus.technion.ac.il}, \ \texttt{\{idogalil.ig,yoad.tewel\}@gmail.com} \\ \texttt{belinkov@technion.ac.il}     \\   
}

\begin{document}
\maketitle
\begin{abstract}
Text-to-image (T2I) diffusion models rely on encoded prompts to guide the image generation process.
Typically, these prompts are extended to a fixed length by appending padding tokens to the input.
Despite being a default practice, the influence of padding tokens on the image generation process has not been investigated.
In this work, we conduct the first in-depth analysis of the role padding tokens play in T2I models.
We develop two causal techniques to analyze how information is encoded in the representation of tokens across different components of the T2I pipeline.
Using these techniques, we investigate when and how padding tokens impact the image generation process.
Our findings reveal three distinct scenarios: padding tokens may affect the model's output during text encoding, during the diffusion process, or be effectively ignored.
Moreover, we identify key relationships between these scenarios and the model's architecture (cross or self-attention) and its training process (frozen or trained text encoder).
These insights contribute to a deeper understanding of the mechanisms of padding tokens, potentially informing future model design and training practices in T2I systems.\footnote{Code available at \href{https://padding-tone.github.io}{padding-tone.github.io}} %

\end{abstract}

\begin{figure}[h!]  %
  \centering
  \includegraphics[width=.48\textwidth]{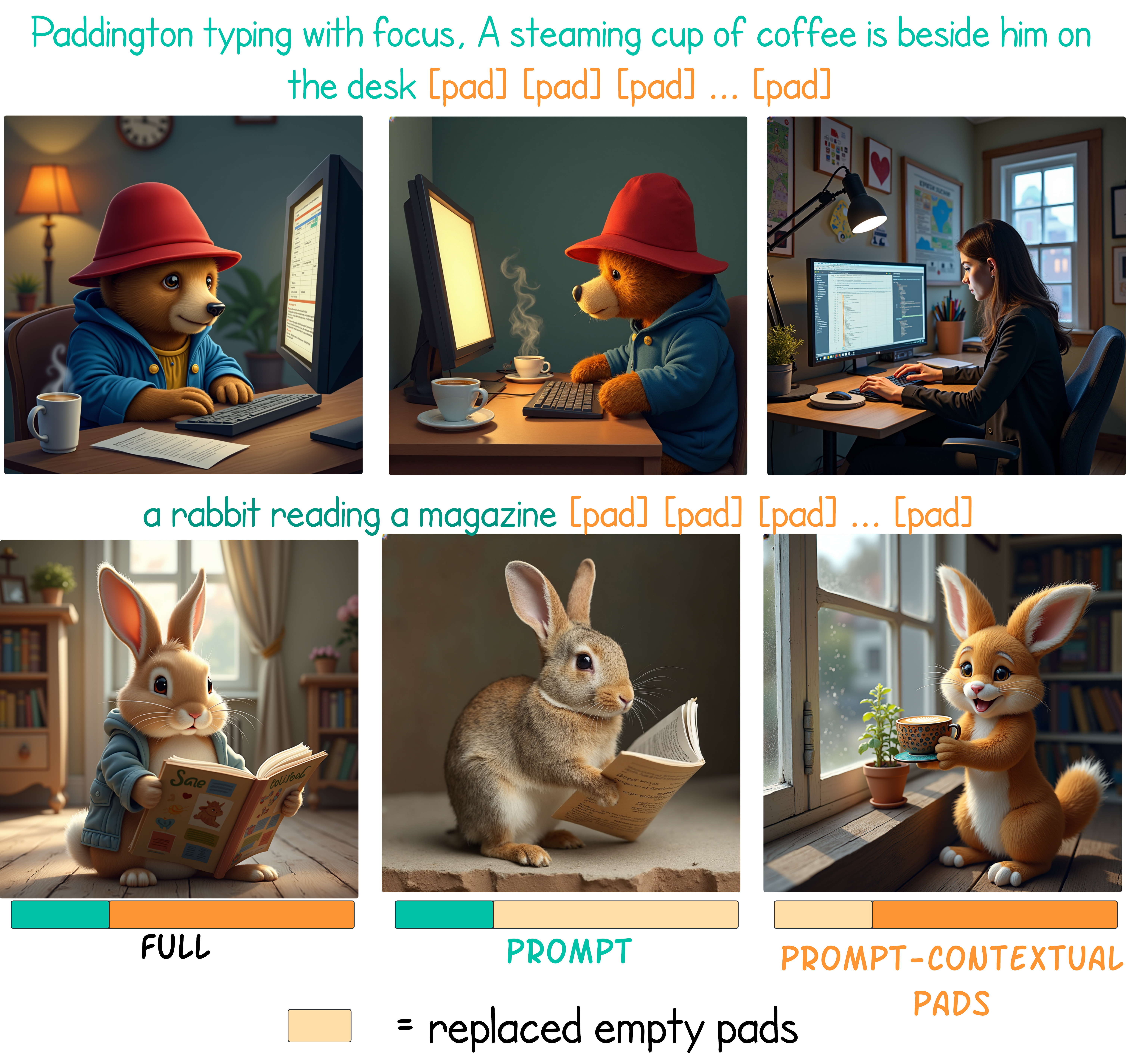}  %
  \caption{Images generated with FLUX from different segments of the input prompt. 
  Description of each column, from left to right: (1) An image generated using the full prompt (both prompt tokens and padding tokens encoded together), (2) An image generated using only the \textcolor{text_tokens}{prompt tokens} and \textcolor{blank_pads}{clean padding tokens}, (3) An image generated using only the \textcolor{prompt_pads}{prompt-contextual pads} encoded with the prompt, while the prompt tokens were replaced with \textcolor{blank_pads}{clean pad tokens}. 
  }
  \label{fig:FLUX_pads}
\end{figure}

\section{Introduction}

Text-to-image (T2I) models consist of two main components: a text encoder, which generates representations of the user’s prompt, and a diffusion model, which generates an image based on this representation.
To standardize sequence lengths for efficient batch processing in training and inference, input prompts are padded to a fixed length with a special padding token.
Unlike language models, where padding tokens are explicitly masked and thus ignored, the computation process of the T2I models can use these tokens as any other token.
Despite their ubiquity, the potential impact of padding tokens on image generation has been overlooked. Our goal in this work is to evaluate the influence of these tokens and determine whether the model learns to use semantically meaningless tokens.

We introduce two methods to evaluate the influence of tokens on different model components: (1) Intervention in the Text Encoder (\encoderMethod) and (2) Intervention in the Diffusion Process (\diffusionMethod).
Both methods build on causal mediation analysis, also known as activation patching \citep{imai2010general, vig2020causal, zhang2024towards}.
This technique involves perturbing specific inputs or intermediate representations to observe their effect on the output, helping to pinpoint the influential elements.
Figure \ref{fig:FLUX_pads} illustrates one of our interventions, showing images generated with perturbations in different parts of the textual representation.

\begin{figure*}[t]  %
  \centering
  \includegraphics[width=\textwidth]{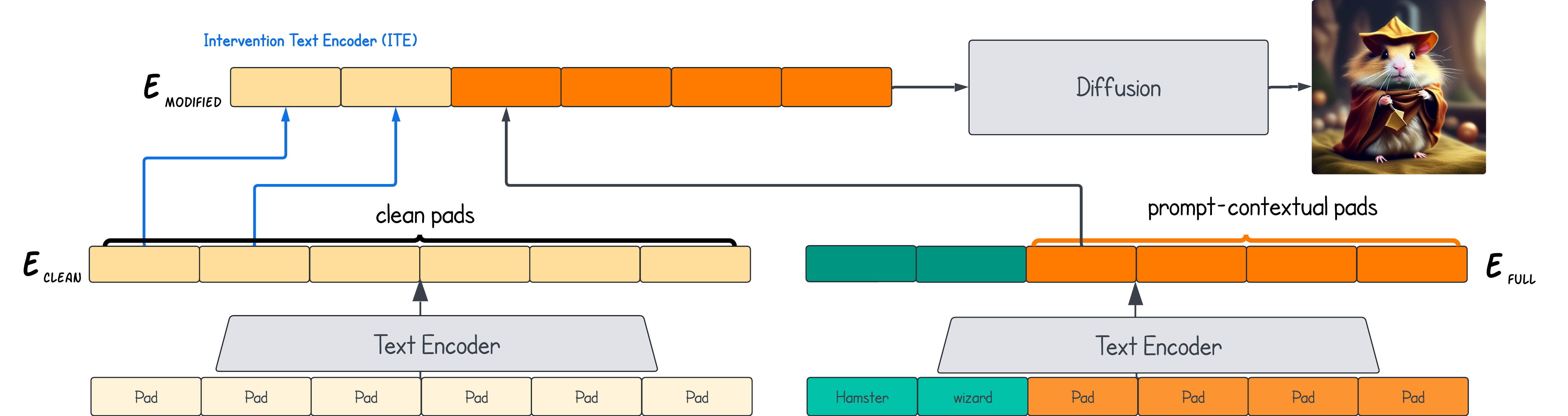}  %
  \caption{\encoderMethod: Interpreting information within pad tokens in the text encoder. We first encode the full prompt and the clean pads separately. Next, we keep the tokens we want to interpret and replace all other tokens with clean pad tokens. We then generate an image conditioned on this mixed representation. In the example shown here, we interpret the pad tokens in LLaMA-UNet, revealing semantic information embedded within the pad tokens.}
  \label{fig:method}
\end{figure*}

In~\encoderMethod~we selectively perturb specific segments of the text encoder’s output representations to isolate the contributions of two key elements: prompt tokens and padding tokens.
Next, we generate images using the modified prompt representations and analyze the results.
The perturbation involves replacing selected token representations with those from a prompt that consists solely of padding tokens, referred to as \textit{clean pads}.
These clean pads differ from the original padding tokens, which contain contextual information from the prompt.
The method is illustrated in Figure \ref{fig:method}.
If padding tokens carry meaningful information, we expect two outcomes: (a) replacing the prompt tokens with clean pads should still result in an image reflecting elements of the original prompt, while (b) replacing the padding tokens with clean pads should alter the image either semantically or stylistically.

In cases where our analysis with~\encoderMethod~indicates that padding tokens are not used by the text encoder, we further examine the role of padding tokens in the diffusion process.
Particularly, we investigate whether significant information is written into the padding token representations throughout the diffusion process.
Here we employ~\diffusionMethod, illustrated in Figure~\ref{fig:diffusion_causal}, to interpret the causal effect of the padding tokens during the diffusion process.
We begin with a standard prompt padded to a fixed length, as well as an ``only pads'' prompt.
However, in~\diffusionMethod, token replacement occurs before each attention block within the diffusion process and at every diffusion step.
We repeat the procedure of selectively replacing either prompt tokens or padding tokens with clean pads, similarly to~\encoderMethod. 

We analyze six different T2I models and highlight two scenarios where padding tokens are utilized.
First, when the text encoder was not frozen during training or fine-tuning, it learns to encode meaningful semantic information into these tokens.
Second, in architectures with multi-modal attention mechanisms---such as Stable Diffusion 3 \citep{esser2024scaling} and FLUX\footnote{\href{https://blackforestlabs.ai/}{blackforestlabs.ai}}---padding tokens carry meaningful information throughout the diffusion process, even if the text encoder itself does not directly encode it.
Here, the padding tokens seem to act as ``registers'', with information written into their representations to store and recall, similarly to findings from both language models and vision-language models  \citep{darcetvision,burtsev2020memory}.
 
\begin{figure}[t]  %
\centering
\includegraphics[width=.48\textwidth]{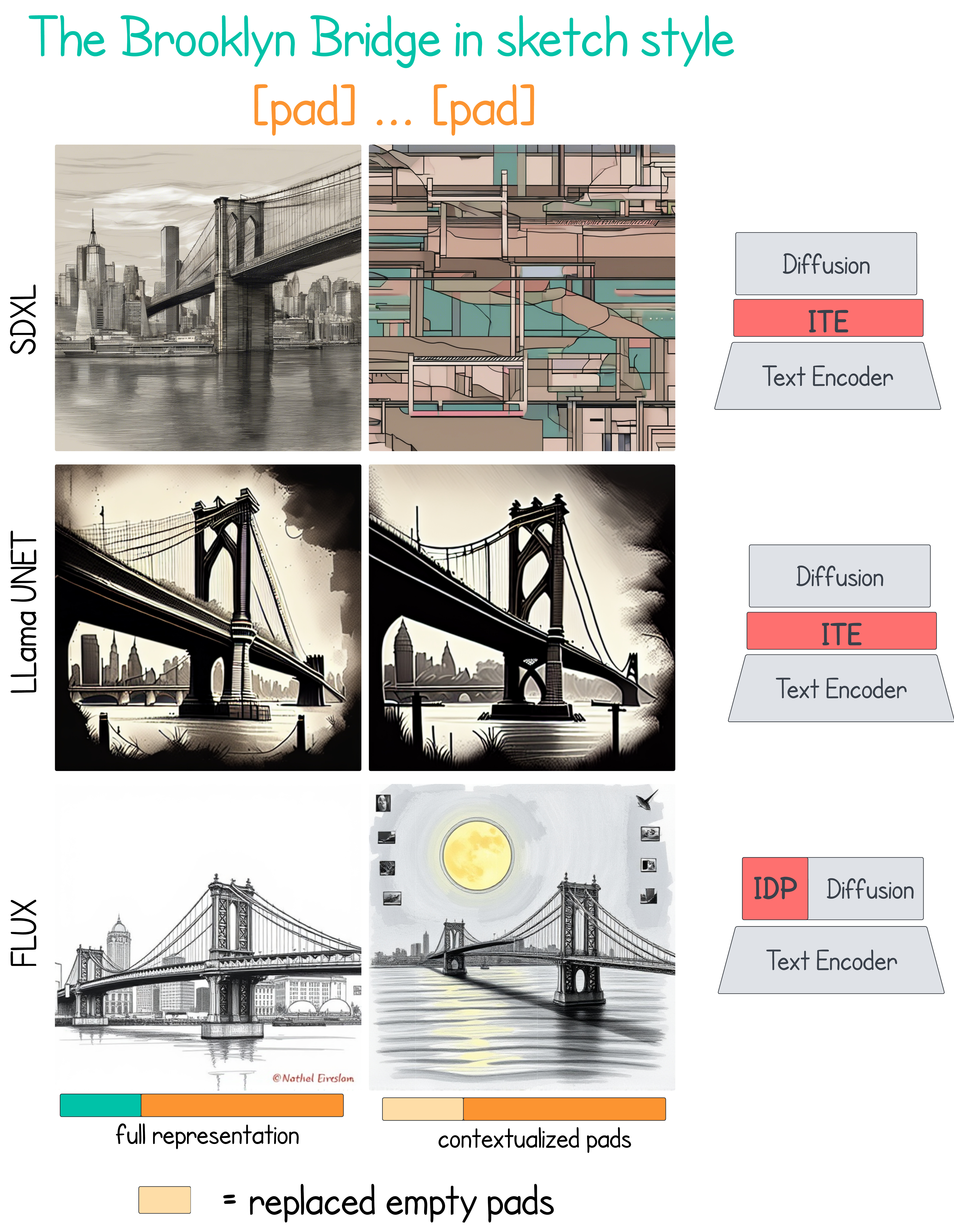}  %
\caption{The scenarios we observe: padding tokens may be effectively ignored (first row; image generated using \encoderMethod), affect the model's output during text encoding (second row; image generated using \encoderMethod), or be used during the diffusion process (last row; image generated using \diffusionMethod). Left: baseline. Right: our method.}
\label{fig:main_scenarious}
\end{figure}

To summarize, our main contributions are:
\begin{enumerate}[itemsep=1pt,parsep=1pt,topsep=1pt]
\item We propose two causal methods for analyzing the use of specific tokens in both the text encoder and diffusion model of the T2I pipeline, and apply them to investigate the role of padding tokens.
\item We find that T2I models with frozen text encoders (e.g., Stable Diffusion XL and Stable Diffusion 2) ignore padding tokens (Figure~\ref{fig:main_scenarious}, first row).
However, when the text encoder is trained or fine-tuned (LDM, LLama-UNet), padding tokens gain semantic significance (Figure~\ref{fig:main_scenarious}, second row).
\item We uncover that even when padding tokens are not utilized by the text encoder, for some architectures with multi-modal self-attention in the diffusion model (Stable Diffusion 3 and FLUX), they can still function as ``registers'' and play a meaningful part in the diffusion process (Figure~\ref{fig:main_scenarious}, last row).
\end{enumerate}

\begin{figure}[t]  %
  \centering
  \includegraphics[width=.48\textwidth]{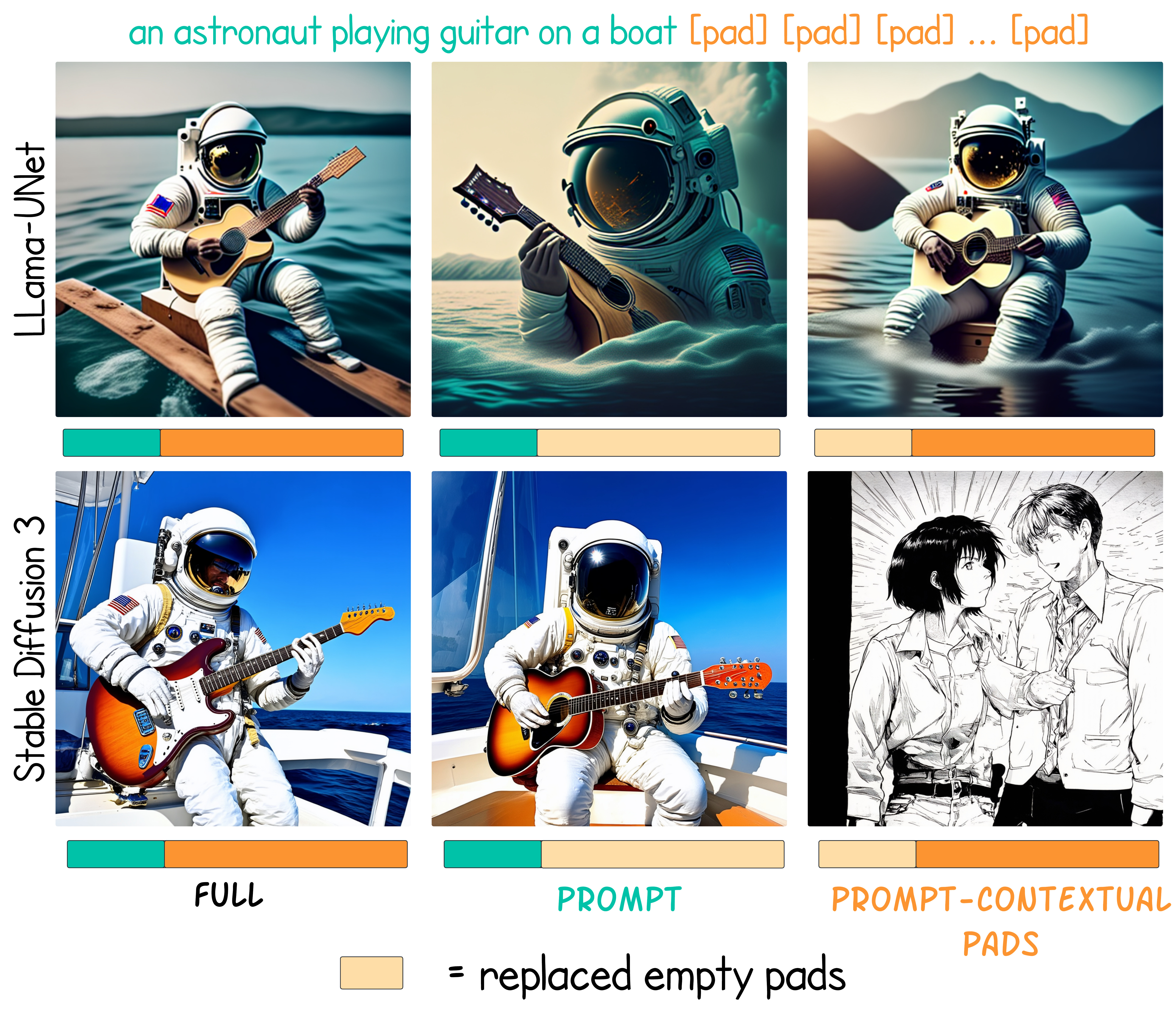}  %
  \caption{Images generated from different segments of the input prompt using~\encoderMethod.   Description of each column, from left to right: (1) An image generated using the full prompt (both prompt tokens and padding tokens encoded together), (2) An image generated using only the \textcolor{text_tokens}{prompt tokens} and \textcolor{blank_pads}{clean padding tokens}, (3) An image generated using only the \textcolor{prompt_pads}{prompt-contextual pads} encoded with the prompt, while the prompt tokens were replaced with \textcolor{blank_pads}{clean pad tokens}. 
}
\label{fig:pads_llama}
\end{figure}

\section{Analysis of Padding in Text Encoding}
\label{sec:method}

In the T2I pipeline, the text encoder processes the input prompt \( P = [P_1, .., P_k] \), \( k \) tokens.
To ensure a consistent input length, the prompt is usually padded to a fixed length, denoted as \( N \).
We denote this padded version of the prompt as \(P_{\text{full}}\), which is a concatenation of the \( k \) prompt tokens and the \( N - k \) padding tokens:
\begin{equation}
\ P_{\text{full}} = [P_1, \dots, P_k, \text{\pad}, \dots, \text{\pad}] .
\end{equation}

The text encoder then processes \( P_{\text{full}} \), producing a constant-length encoded representation, which is subsequently used by the diffusion model for conditional image generation.
We denote this encoded full prompt representation as \( E_{\text{full}} \), which consists of the encoded prompt tokens and the encoded prompt-contextual padding tokens.\footnote{One special case is the use of an EOS token in CLIP models, which is discussed in Appendix~\ref{app:eos_clip}.}

\subsection{Method}
Our goal is to evaluate the information encoded in the prompt-contextual padding tokens, and to measure their effect on the generated image.
To do so, as illustrated in Figure~\ref{fig:method}, we generate images using partial representations of \( E_{\text{full}} \) that isolate the effect of the padding tokens.
We generate images based on these partial representations of \( E_{\text{full}} \) and compare them to images generated from the full prompt \( E_{\text{full}} \). 
This enables us to visually express the information from different parts of the text input.

Specifically, to remove information coming from a subset of the tokens, we replace them with ``clean'' padding tokens that were not influenced by the user's prompt.
To obtain these clean padding tokens, we encode \( S_{\text{clean}} =  [\text{\pad}, \text{\pad}, \dots ,\text{\pad}] \), a fixed-length sequence made entirely of padding tokens, and denote their embeddings as $E_{\text{clean}}$.

These encoded padding tokens are then used in constructing the final mixed representation, which combines both the prompt-contextual tokens and clean padding tokens.
We use the encoded padding tokens since they contain no information related to the current prompt, while maintaining the same length and distribution of the text encoder's output.
This allows us to effectively isolate the contribution of the padding tokens that are encoded alongside the full prompt tokens, helping us understand how much of the information in the final representation comes from the prompt itself versus the prompt-contextual padding tokens.
Figure \ref{fig:pads_llama} demonstrated our method.
First, we generate an image from the full prompt, which is how the image is generated in the standard pipeline (left column).
Then, we generate an image that demonstrates the information in the non-pad tokens (e.g. encoded prompt tokens), by replacing the prompt-contextual padding tokens with clean pads (middle column).
Lastly, we generate an image demonstrating the information within the prompt-contextual padding tokens, by replacing the non-pad tokens with clean pads (right column).

More formally, the mixed representation for generating an image from the encoded prompt tokens only (middle column) is:

\begin{equation}
\label{eq:prompt}
    E_{\text{prompt}} = \left[ E_{\text{full}}^{0:k}, E_{\text{clean}}^{k+1:N-1} \right],
\end{equation}
where \( E_{x}^{i:j} \) represents the encoded tokens from index $i$ to $j$. For a representation that generates an image from the prompt-contextual padding tokens only (right column):

\begin{equation}
\label{eq:pads}
    E_{\text{pads}} = \left[ E_{\text{clean}}^{0:k}, E_{\text{full}}^{k+1:N-1} \right] 
\end{equation}

\begin{figure*}[t]  %
  \centering
  \includegraphics[width=\textwidth]{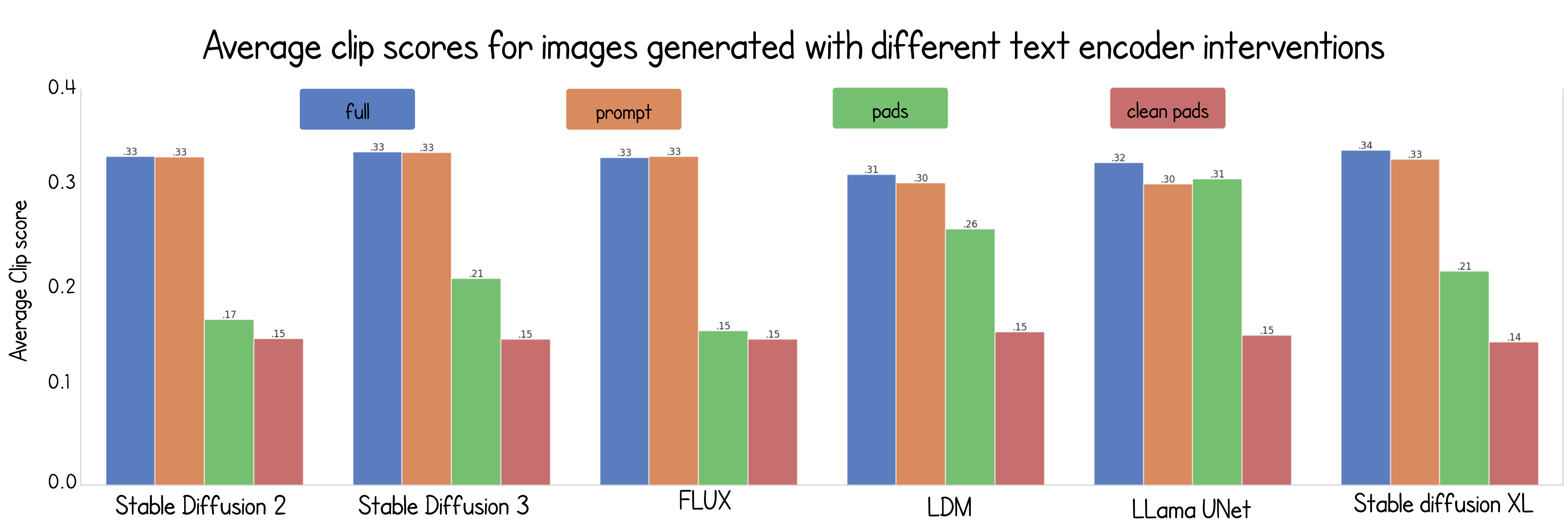}  %
    \caption{
  Average CLIP score over 5,000 images generated from the different representations: \textcolor{blue_fig_4}{full prompt}, \textcolor{orange_fig_4}{only prompt}, \textcolor{green_fig_4}{prompt-contextual pads} and \textcolor{red_fig_4}{clean pads} using \encoderMethod. LDM and LLaMA-UNet are the only models achieving high CLIP scores for images generated from padding tokens, indicating their use during text encoding. See Table \ref{app:tab:main_std} in the Appendix for  standard deviations.}
  \label{fig:main_results}
\end{figure*}

\subsection{Experimental Setup}
\paragraph{Models.}
\label{sec:text_encoder:experimental}
We use six T2I models.
These models can be divided into two categories based on their training approach: 
those with pretrained frozen text encoders during the training: Stable Diffusion 3 \citep{esser2024scaling}, Stable Diffusion 2, Stable Diffusion XL \citep{DBLP:conf/iclr/PodellELBDMPR24}, FLUX; and those with some learned weights as part of the text to image training: LDM \citep{DBLP:conf/cvpr/RombachBLEO22} and Lavi-Bridge \citep{zhao2024bridging} (LLaMA-UNet version).
The first group can be divided to two subgroups:
models that use vision-language cross-attention with the text representations in the diffusion process   (Stable Diffusion 2, Stable Diffusion XL) and models that use the text representations as part of vision-language self-attention, allowing text representations to change throughout diffusion (FLUX, Stable Diffusion 3).
Appendix~\ref{app:models} provides more information regarding each of the models.

\paragraph{Data.} Our prompts are based on the Parti dataset \citep{Yu2022ScalingAM}, a benchmark containing over 1600 diverse and challenging prompts used to evaluate T2I models.
To prevent using prompts that have leaked into the training corpus of the models, we select prompts from eight different challenge categories in Parti, and use GPT-4o\footnote{\href{https://openai.com/index/hello-gpt-4o/}{openai.com/index/hello-gpt-4o}} to generate an alternative set of prompts with similar style and complexity.
We then manually review the prompts to ensure their coherence.
This process results in 500 new prompts.
The complete list of categories, along with the prompt used with GPT, can be found in Appendix \ref{app:data_creating}, and the full dataset is included in the supplementary material.

Each of the 500 prompts is used to generate 10 images from different random seeds, resulting in 5,000 images for each configuration of model and representation.
We investigate three representations: $E_\text{full}$, $E_{\text{prompt}}$ (Eq.~\ref{eq:prompt}), $E_{\text{pads}}$ (Eq.~\ref{eq:pads}), and $E_{\text{clean}}$ as a lower-bound control, with their corresponding images denoted as ``full'', ``prompt'', ``prompt-contextual pads'' and ``clean'', respectively. 

\paragraph{Metrics.} To evaluate the generated images, we employ two key metrics: CLIP score \citep{hessel-etal-2021-clipscore}, which  measures how well the generated images align with the textual prompt,
and KID (Kernel Inception Distance) \citep{binkowski2018demystifying}, to evaluate the quality of generated images. KID is used 
to measure the similarity between the feature distributions of images generated from full representation and images generated after some causal intervention. Unlike FID \citep{NIPS2017_8a1d6947}, which is based on Gaussian approximations, KID uses the maximum mean discrepancy (MMD) measure, making it more robust in practice, especially when dealing with smaller sample sizes.

\subsection{Results}

Figure \ref{fig:main_results} shows the average CLIP scores over generations from different representations: ``full'', ``prompt'', ``prompt-contextual pads'' and ``clean''.
Stable Diffusion (versions 2+3) and FLUX models appear to make little to no use of padding tokens: CLIP scores for the full and prompt representations are nearly identical, while the prompt-contextual pads—containing only padding tokens—yield significantly lower scores.
In contrast, LLaMA UNet and LDM  contain significant semantic information in padding, with a higher CLIP score for the ``prompt-contextual pads'', although the degradation in performance from ``full'' to ``prompt'' is small.

\begin{table}[t]
\centering
\begin{tabular}{l cr }
\toprule 
& \multicolumn{2}{c}{\textbf{KID Score}} \\ 
\cmidrule(lr){2-3}
                                 \textbf{Model} & \textbf{Prompt}  & \textbf{Pads}        \\ 
                                \midrule 
\textbf{FLUX}             & $0.01$             & $14.52$                \\ 
\textbf{LDM}                      & $0.88$             & \underline{$4.53$}                 \\ 
\textbf{LLaMA UNet}               & $7.37$             & \underline{$0.48$}                 \\ 
\textbf{Stable Diffusion 2}       & $0.02$             & $31.09$                \\ 
\textbf{Stable Diffusion 3}       & $0.01$                & $15.74$                \\ 
\bottomrule 
\end{tabular}
\caption{KID scores between the images generated from the prompt-contextual pads vs.\ images generated only from prompt representations. All images are generated using \encoderMethod. Lower is better. The KID is calculated w.r.t. images generated from the full representation. Notably, LDM and LLaMA UNet are the only models that achieve a low KID on images generated from contextual pads (\underline{underlined}).}
\label{tab:kid}
\end{table}

\paragraph{Text encoder training objective and its influence on padding usage.}
Our results suggest that the training objective of the text encoder significantly impacts how padding tokens are utilized. 
Many current T2I models, such as Stable Diffusion and FLUX, employ a frozen text encoder, with the diffusion model being trained on its encoded outputs.
It may be that because the text encoder is not explicitly trained to process padding tokens for image generation, it does not effectively incorporate them during the textual encoding.
As shown in Figure~\ref{fig:main_results}, in models that use frozen text encoders (Stable Diffusion and FLUX), images generated using the ``prompt'' representation yield the same CLIP score as those generated using the ``full'' representation, while images generated from ``prompt-contextual padding'' representations result in a very low CLIP score, almost as low as those generated from clean padding. Furthermore,
Table~\ref{tab:kid} shows a clear distinction between models trained to process padding tokens for image generation and those that are frozen. For models with a frozen text encoder, the KID is very low (around 0.01) for images generated from prompt tokens only, whereas it is high for images generated from contextual padding tokens, indicating that these images are out of distribution. This suggests that in these models, the text encoder does not encode meaningful information in the padding tokens, making them unnecessary for generating the final image.

Other models, like LDM and Lavi-Bridge, adapt the text encoder specifically for the image generation task. These methods train the text encoder, including the use of padding tokens, on the image generation objective, allowing it to effectively learn how to utilize padding. In these models, the results differ: images generated from full prompt tokens have lower scores compared to those generated using prompt representations, suggesting that the information encoded in the prompt tokens is insufficient to generate the precise images.
Furthermore, images generated from the prompt-contextual padding tokens in these models yield much higher CLIP scores, even surpassing images generated from prompt tokens in LLama-UNet. These models achieve relatively low KID on images generated from the prompt-contextual padding tokens, indicating that these images are close in distribution to those generated from the full representation. This suggests that in these models, the text encoder has learned to utilize the padding tokens during the textual encoding.

Overall, these results indicate that pads play an important role in the text encoding process for image generation in these adapted models.
\paragraph{How many padding tokens do text encoders use?} 
We focus on the LLaMA-UNet model and analyze padding behavior. We divide the padding tokens into five segments, each containing 20\% of the total padding tokens in their natural order. For each segment, we mask both the prompt tokens and pad tokens in the other segments, then generate images from this mixed representation.

\begin{table}[t]
\centering
\begin{tabular}{l r}
\toprule
\textbf{Pad Segment} & \textbf{ CLIP Score} \\ \midrule
  First        &  $0.30$ \small{$\pm 0.018$}                \\
  Second        &  $0.23$ \small{$\pm 0.018$}                \\
  Third        &  $0.17$ \small{$\pm 0.022$}                \\ \bottomrule
\end{tabular}%
\caption{Average CLIP scores for different prompt-contextual pad segments in LLaMA-UNet: the first 20\% of the pads, the next 20\%, and then the subsequent 20\%. We observe that the semantic information degrades gradually, with most of it concentrated in the initial tokens.}
\label{tab:pad_segments}
\end{table}

The CLIP scores can be found in Table \ref{tab:pad_segments}. Our observations reveal that the information encoded in padding tokens varies based on their proximity to the prompt tokens, with those closer to the prompt carrying more significant information. We hypothesize that this behavior may be due to the text encoder's use of causal masking or the positional encoding scheme applied to the padding tokens. Only the padding tokens that are closer to the prompt tokens appear to be utilized effectively.

\begin{figure}[t]  %
  \centering
  \includegraphics[width=.48\textwidth]{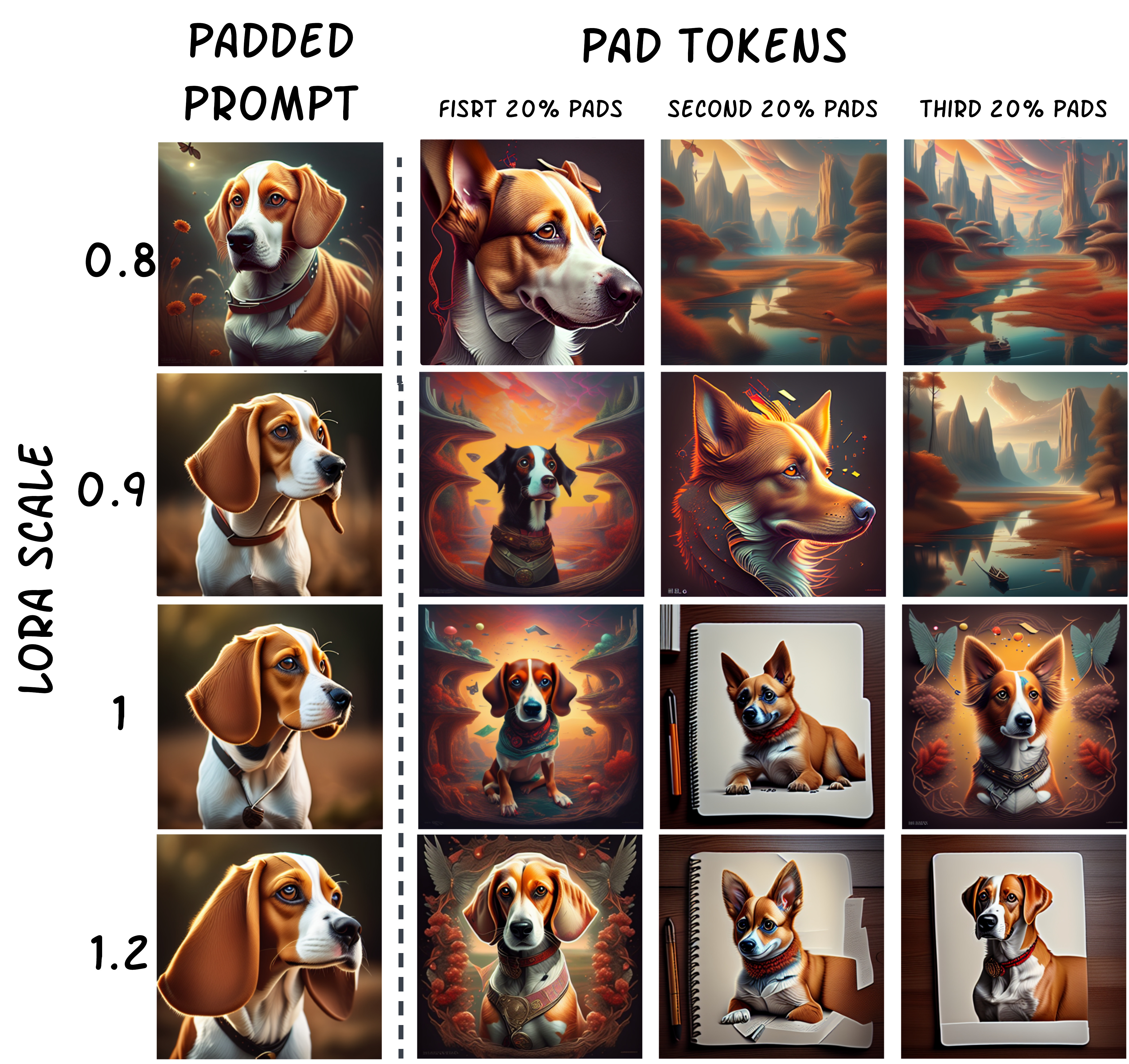}
  \caption{Images generated from Lavi-bridge with LoRa loaded with scaling factor \(\alpha\) (y-axis). We analyze pad token segments: the first column shows the full image, and the next columns show three consecutive 20\% of the pads. As \(\alpha\) decreases, fewer pad tokens are used.
}
  \label{fig:lora}
\end{figure}

Since LLaMA is a language model adapted for image generation using LoRa training, we can load the LoRa with a scaling factor, \(\alpha\), to observe how gradually removing LoRa affects the number of used pad tokens. Our results in Figure~\ref{fig:lora} show that as \(\alpha\) decreases, fewer pad tokens are used. This indicates that part of what the LoRa learns involves encoding information into more pad tokens.

\begin{figure}[t]  %
  \centering
  \includegraphics[width=0.48\textwidth]{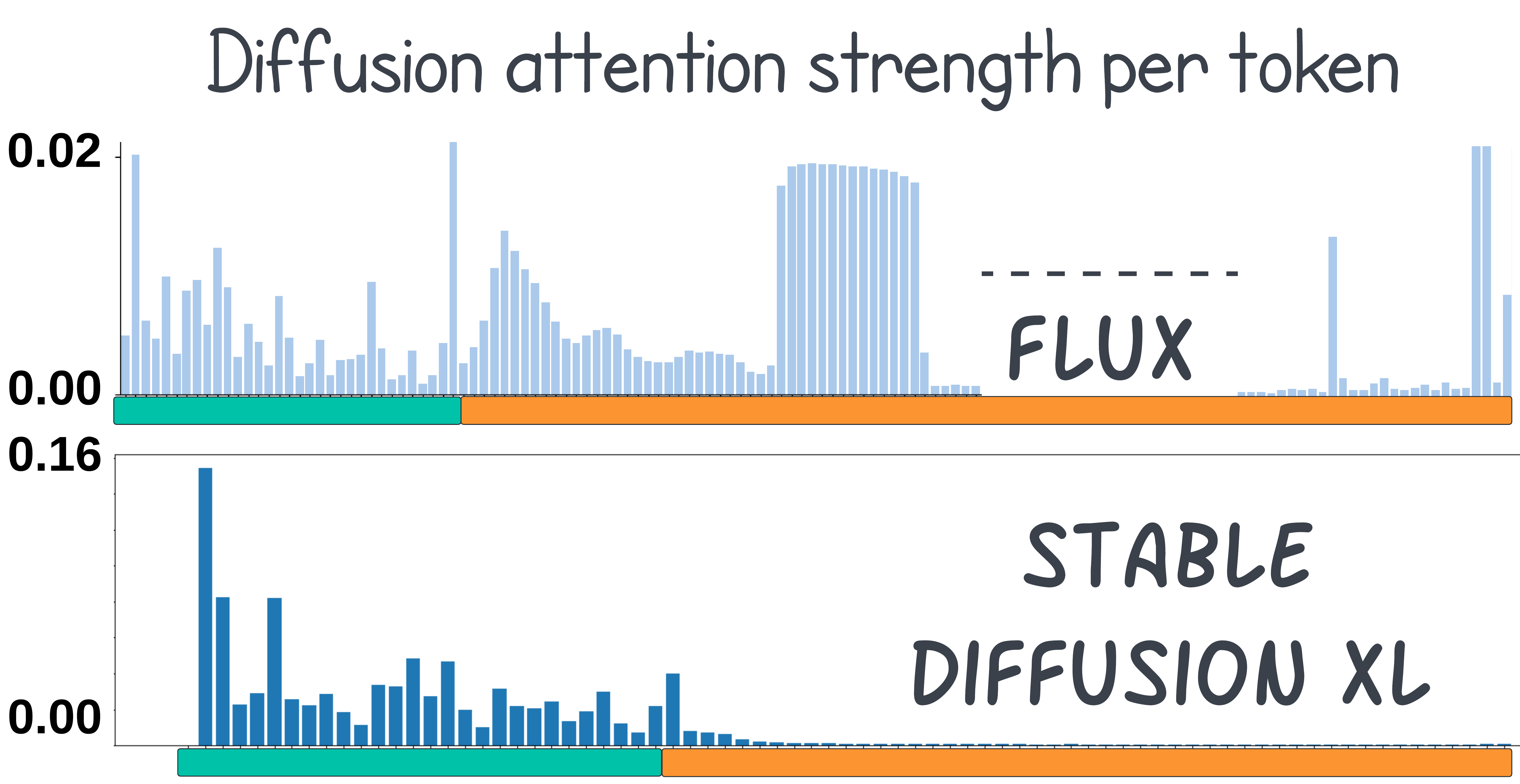}  %
  \caption{Attention histogram for Stable Diffusion XL and FLUX* for each token reveals that while both models exclude semantic information from padding tokens, FLUX utilizes these tokens, whereas Stable Diffusion does not. *In FLUX, we have removed the long middle part with low attention in order to improve visualization.}
  \label{fig:ca_strength}
\end{figure}

\begin{figure}[]  %
  \centering
  \includegraphics[width=.49\textwidth]{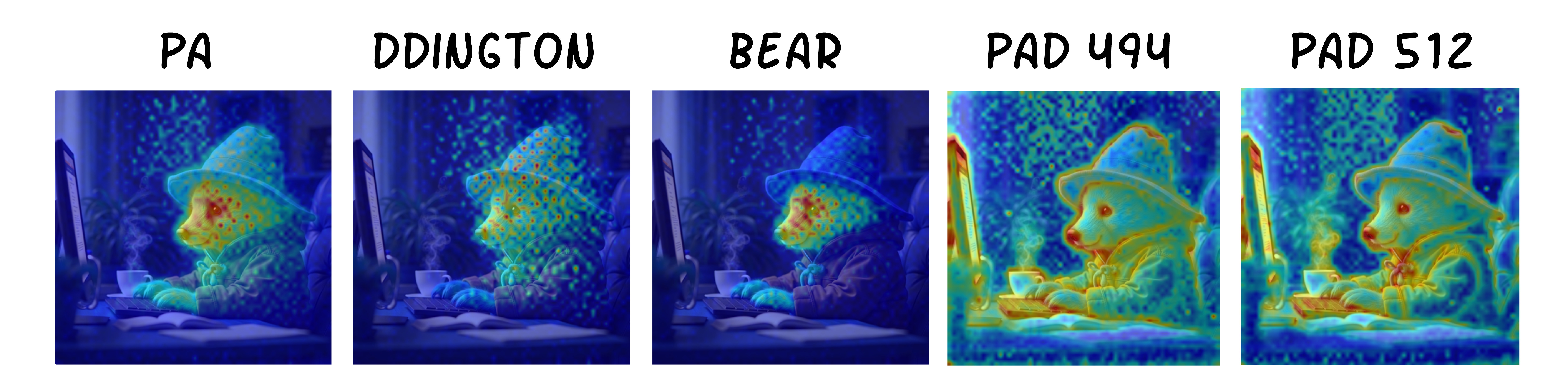}
  \caption{Attention maps for FLUX diffusion show strong alignment between prompt tokens and semantically relevant image tokens. These maps also reveal high attention for padding tokens with the main objects in the image.
  }
  \label{fig:token_attention_maps}
\end{figure}

\begin{figure*}[t]  %
  \centering
  \includegraphics[width=\textwidth]{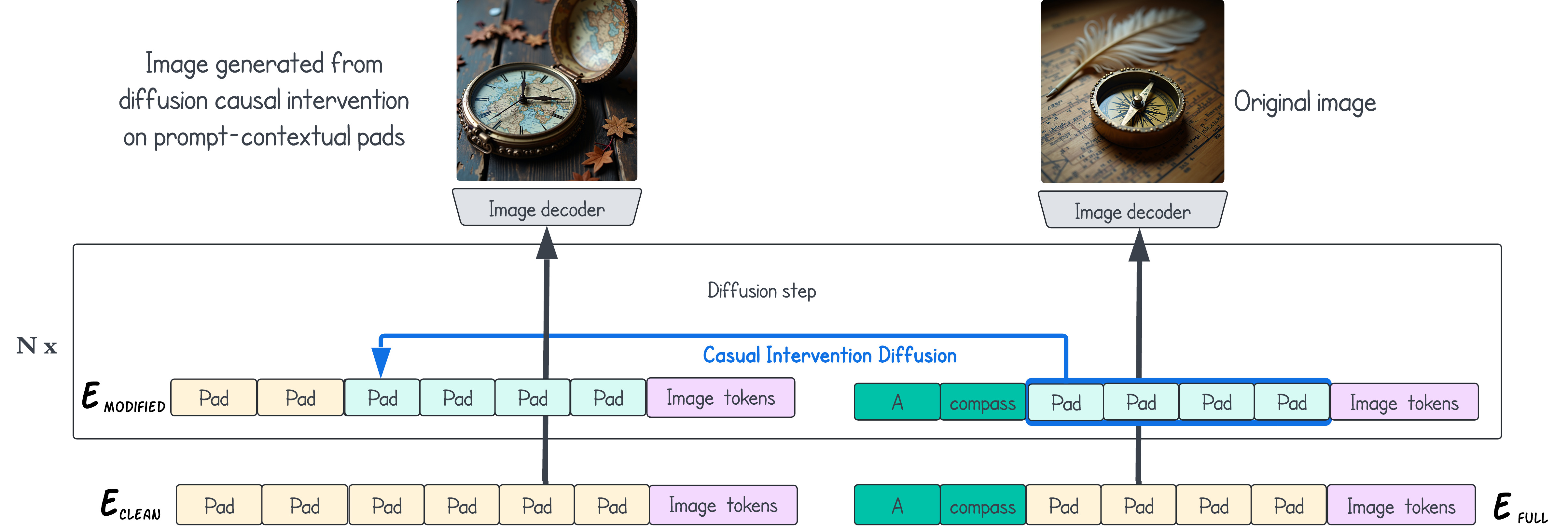}  %
  \caption{\diffusionMethod: Interpreting information within pad tokens in the diffusion model. We perform a diffusion of two prompts simultaneously: the full prompt and an clean pads. During the diffusion, we keep the tokens we want to interpret (here:  the prompt-contextual padding tokens) and replace all other tokens with clean pad tokens. We perform this intervention before each attention block in the diffusion model, through all diffusion steps. We then generate an image conditioned on this mixed representation. In the example shown here, we interpret the pad tokens in FLUX, revealing semantic information embedded within the pad tokens during diffusion.}
  \label{fig:diffusion_causal}
\end{figure*}

\section{Analysis of Padding in the Diffusion Process}
\label{analysis_diffusion}
Padding tokens may not carry meaningful information after text encoding, yet some diffusion architectures may still utilize them during the diffusion process.
To generate images from the encoded textual representation, T2I models use an attention mechanism to condition the generation process. This mechanism typically follows one of two common approaches: cross-attention and MM-DiT~\citep{esser2024scaling} blocks.
In cross-attention, used in models like Stable Diffusion 2/XL, the model converts image patches into query vectors and text tokens into key and value vectors.
The image patches gather information from the encoded textual representation, based on an attention map, but the text representation remains unchanged throughout the process.
In contrast, MM-DiT blocks, found in models like FLUX and Stable Diffusion 3, implement a multi-modal self-attention,  by projecting both image patches and text token representations into query, key, and value vectors.
Thus, both the image and text representations update and influence each other during the attention process.
We therefore expect that models implementing cross-attention where the pads are not used in the text encoder would also not use them in the diffusion process.
However, models implementing \textit{MM-DiT} blocks can potentially aggregate information into the padding tokens, even if initially they contain no information.

\paragraph{Motivation: attention maps and qualitative examples.}
To understand whether padding tokens are utilized during image generation, we examine the attention maps between image patches and text representations, resulting in an attention map between each text token and all the image tokens (see example in Figure~\ref{fig:token_attention_maps}).
While in Stable Diffusion XL only the prompt (and the end-of-text) tokens significantly attend to main areas in the image, in FLUX not only prompt tokens, but also many pad tokens contribute much attention to main image areas (Figure \ref{fig:ca_strength}). Moreover, generating images with FLUX and Stable Diffusion XL, with and without padding (Figure \ref{fig:max_len}, Appendix~\ref{app:qual}), reveals that FLUX without padding often misses key details, while Stable Diffusion XL remains consistent in its generations even without padding tokens.

\subsection{Method}
To interpret the causal effect of tokens during the diffusion process, we develop~\diffusionMethod, illustrated in Figure \ref{fig:diffusion_causal}. The diffusion process consists of several diffusion steps, where each step begins with the current latent image representation and the full encoded text representation. Since we look only at models where padding tokens do not carry meaningful information in the text encoder, we hypothesize that the diffusion model might be using these tokens as ``registers'' to store and recall information, subsequently passing it to the image patches, similar to the findings of \citet{darcetvision} in their work on VLMs with image patches.

To investigate the role of padding tokens during the diffusion process, we intervene before each attention block at every diffusion step.
More specifically, we use two representations: the fully encoded padded prompt representation from the text encoder and another encoded ``clean pads'' prompt, whose representations per diffusion layer are denoted as $E^{(l)}_{\text{full}}$ and $E^{(l)}_{\text{clean}}$, respectively. %
We replace the prompt tokens with clean pads at each diffusion step, before each transformer layer using Equation~\ref{eq:prompt}, resulting in an image generated solely through self-attention to the prompt-contextual pads.
If the images generated from these representations still contain prompt-relevant information, it would suggest that the pads are being utilized during the diffusion process.

\begin{table}[t]
\centering
\begin{tabular}{l r r }
\toprule
& \multicolumn{2}{c}{\textbf{CLIP Reference}} \\ 
\cmidrule(lr){2-3}
\textbf{Representation} & \textbf{Image} & \textbf{Prompt} \\ \midrule
Full   & $1.0$  \small{$\pm 0.0$}    & $0.33$  \small{$\pm 0.037$} \\ 
Prompt & $0.91$  \small{$\pm 0.003$} & $0.33$  \small{$\pm 0.038$} \\
Pads   & $0.72$  \small{$\pm 0.008$} & $0.22$  \small{$\pm 0.054$} \\
Pads First   & $0.58$  \small{$\pm 0.008$} & $0.21$  \small{$\pm 0.052$} \\
Pads Second  & $0.55$  \small{$\pm 0.006$} & $0.15$  \small{$\pm 0.038$} \\
Clean  & $0.46$  \small{$\pm 0.018$} & $0.10$  \small{$\pm 0.009$} \\
\bottomrule
\end{tabular}%
\caption{Average CLIP scores between images generated  (with FLUX) with different~\diffusionMethod~interventions and either the full prompt or an image generated from the full prompt. `Full': a prompt with real tokens and pads. `Prompt': prompt tokens; `Pad':  prompt-contextual pads; `Pad First': First 20\% of the prompt-contextual pads; `Pad Second': Second 20\% of the prompt-contextual pads;  `Clean':  a prompt full of pads, used for comparison; }
\label{tab:cid_images_all}
\end{table}

\subsection{Results}
Table \ref{tab:cid_images_all} shows the results of our intervention in the diffusion process. The table shows average CLIP scores of images generated with different~\diffusionMethod~interventions, to assess the role of pad and prompt tokens. 
First, we compute CLIP scores with regards to the full-text prompt (Prompt column). We find that images generated only from the prompt tokens are similar to the prompt to the same extent as images generated from the full prompt. Images generated only from the prompt-contextual pads are much more similar to the text prompt compared to images generated from clean pads, indicating that pad tokens are used by the diffusion model to represent concepts related to the prompt. That said, the CLIP scores of images generated from the prompt-contextual pads are much lower compared to the CLIP scores of images generated from prompt-contextual pads in LLama-UNet with~\encoderMethod~. 

\begin{figure}[th!]  %
  \centering
  \includegraphics[width=.48\textwidth]{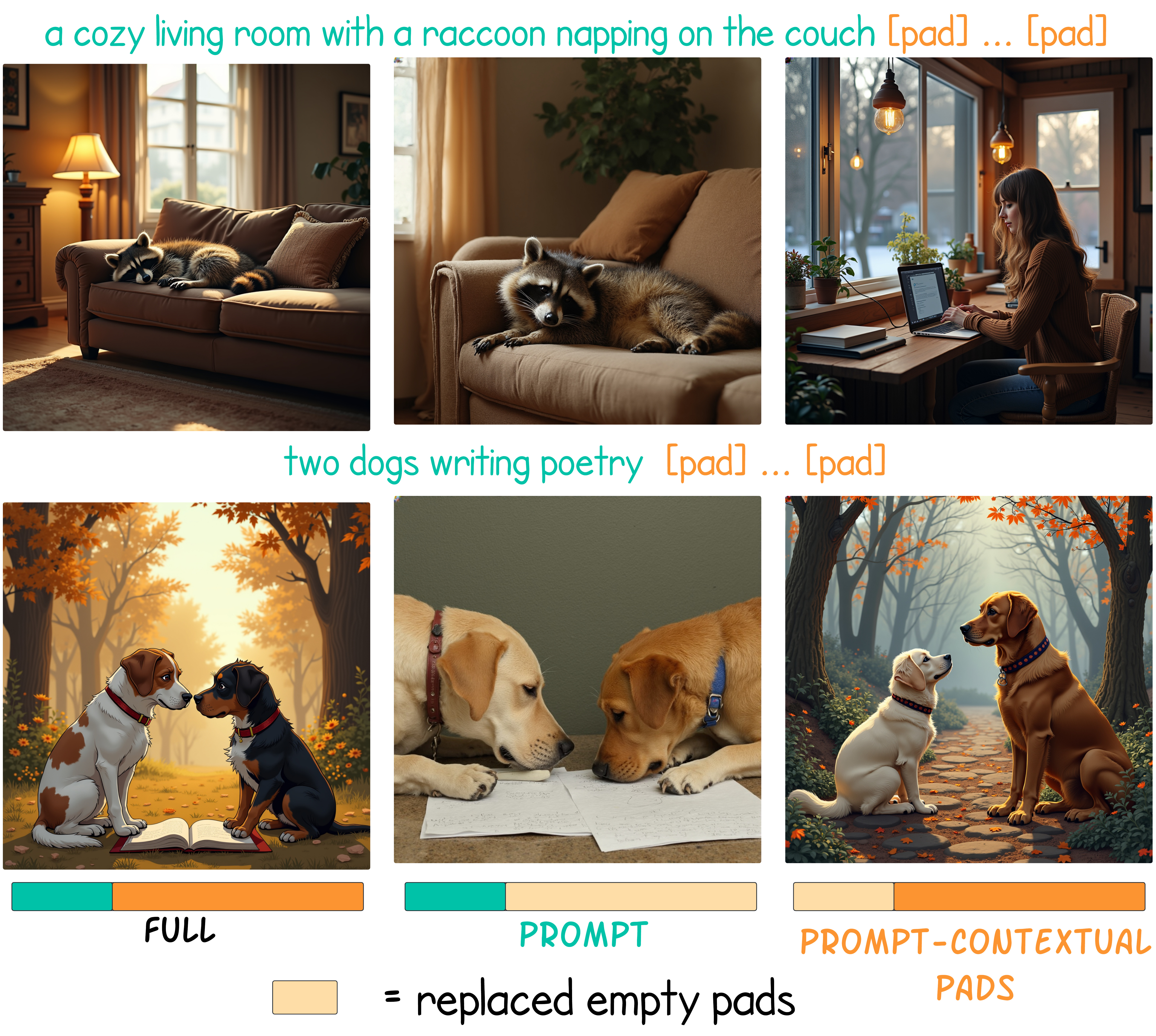}  %
  \caption{
    Images generated with FLUX using~\diffusionMethod~from different prompt segments show distinct alignments: prompt tokens produce semantically accurate images, while the visual nuance like 'cozy' emerges only from the prompt-contextual pad tokens.
  }
  \label{fig:sid_example}
\end{figure}

These results are further demonstrated in Figure~\ref{fig:sid_example}, where we present images generated by ~\diffusionMethod~. In the first row, images are generated for the prompt ``\textit{a cozy living room with a raccoon napping on the couch}''. The image generated from the prompt-contextual pads has a cozy style but does not align well with the prompt itself. In the second row, for the prompt ``\textit{two dogs writing poetry}'', the image generated from the prompt-contextual padding tokens retain several visual features present but do not incorporate the concept of writing poetry, which is prominent in the prompt. We hypothesize that in diffusion, pads are used to represent visually related information but do not contain the same semantic information. 

To further explore this relationship, we compute CLIP scores with regards to images generated from the full padded prompts (Table~\ref{tab:cid_images_all}, Image column).
The CLIP score between images generated from full prompts and images generated when using only contextual padding tokens is approximately $72$---significantly higher than the score for randomly generated images from a `clean' padding prompt. This is further evidence that the padding tokens contain visual information closely related to the content of the prompt tokens. Here, similar to the results in Section~\ref{sec:method}, images generated from the first 20\% of the contextual pads contain more information than those generated from the next 20\%, indicating that information is not spread evenly and is primarily concentrated in the first contextual pads.

Finally, we provide more qualitative examples in Figure \ref{fig:causal_diffusion_examples} (Appendix \ref{app:qual}), which show that images generated from the prompt-contextual pads with~\diffusionMethod~have meaningful semantic information. While images generated solely from prompt tokens typically align with the semantic meaning of the prompt, different visual features are often missing when padding tokens are excluded, while the same features are presented in the padding tokens. It appears that the diffusion model uses padding tokens to create additional visual information, while semantic content remains primarily in the prompt tokens.

\section{Related Work}

\paragraph{Special tokens and additional computation.}
While padding tokens are generally used for efficient batch processing without fulfilling a functional role, other special tokens are known to carry various roles. In transformer language models, attention is often directed to special tokens, including punctuation marks (`.'), [SEP], or just the first token; this has been referred to as null or no-op attention \cite{vig-belinkov-2019-analyzing,kovaleva-etal-2019-revealing,clark-etal-2019-bert,rogers-etal-2020-primer}.
Some have added special tokens to enable additional processing, such as `registers' in vision transformers \cite{darcetvision} or `memory tokens' in language models \cite{burtsev2020memory}.
More generally, language models benefit from additional computation via chain-of-thought reasoning \cite{10.5555/3600270.3602070}.
Finally, several studies found it useful to \emph{train} models to perform additional computation with custom tokens, including  filler tokens like `.....' \cite{pfau2024let}, so-called `pause tokens'  \cite{goyal2024think}, or `meta-tokens' for additional reasoning steps \cite{zelikman2024quietstar}. 
This idea can be traced back to adaptive computation time techniques \cite{graves2016adaptive,banino2021pondernet}.
Our work contributes to this literature by analyzing the role of padding tokens in T2I models. 

\paragraph{Interpreting vision-language models.}
Compared to uni-modal models, VLMs have seen relatively few attempts at interpretation. CLIP~\citep{clip2021} has been a focus of several studies: \citet{goh2021multimodal} identified multimodal neurons responding to specific concepts, while \citet{gandelsman2023interpreting} decomposed its image representations into text-based characteristics. In the realm of text-to-image models, \citet{tang2022daam} introduced a method to interpret T2I pipelines by analyzing the influence of input words on generated images through cross-attention layers. \citet{cheferhidden}  decomposed textual concepts, with a focus on the diffusion component. \citet{basu2024localizing} employed causal tracing to investigate the storage of knowledge in T2I models like Stable Diffusion. \citet{DBLP:conf/acl/TokerOVAB24} analyzed the text encoder in T2I pipelines, offering a view into intermediate representations rather than just its final output.

Our work takes a unique direction by focusing specifically on padding tokens, which have been largely overlooked in prior research.
While previous research has illuminated how prompt tokens guide image generation, we show that padding tokens, often thought to be inert, can play a more active role—encoding semantic information or even functioning as ``registers'' that influence model computations. This adds a new dimension to the interpretation of T2I models, suggesting that even these seemingly unimportant tokens may hold valuable information or operational significance.

\section{Discussion} This work addresses a design decision present in every T2I model that has remained largely unexplored: the choice to include padding tokens during both textual encoding and the diffusion process.
As more studies begin integrating large language models (LLMs) into T2I pipelines using techniques like fine-tuning, LoRA, or adapters, the role of padding tokens becomes increasingly crucial. Training these models with padding tokens could influence a wide range of methods that assume subject information is encoded in specific tokens~\citep{chefer2023attend,rassin2023linguistic,hertz2023prompttoprompt,gal2022image}, potentially altering their implementation when padding tokens carry significant semantic information. This factor should be carefully considered when deciding whether to train with or ignore padding tokens. 

Furthermore, future research could explore how incorporating padding tokens into training might provide computational advantages in more integrated, end-to-end architectures, potentially allowing models to dynamically allocate resources by adjusting the use of padding tokens as needed.

\section*{Limitations}
While we have studied multiple T2I models representing several architectures,  our work did not cover the vast space in this area. Our prompt selection offers some variety, but it may not capture all edge cases, potentially overlooking cases where padding tokens are used differently. Additionally, although we rely on widely used metrics like CLIP score and KID for evaluation, these may not capture all nuances of image quality.

\section*{Ethical Considerations}
In developing our code, we used both Copilot and GPT-4o, but carefully reviewed each line to ensure it aligned with our intended implementation. For writing and rephrasing improvements, we used WordTune and GPT-4o. Every generated suggestion was carefully reviewed and adjusted to ensure our original intent remained intact.

\section*{Acknowledgments}
This research was supported by the Israel Science Foundation (grant 448/20), an Azrieli Foundation Early Career Faculty Fellowship, an AI Alignment grant from Open Philanthropy, and an Academic Gift Award from NVIDIA. HO is supported by the Apple AIML PhD fellowship. The authors would like to thank Mor Ventura for her insightful remarks and valuable suggestions. This research was funded by the European Union (ERC, Control-LM, 101165402). Views and opinions expressed are however those of the author(s) only and do not necessarily reflect those of the European Union or the European Research Council Executive Agency. Neither the European Union nor the granting authority can be held responsible for them.

\bibliography{custom}

\appendix
\label{sec:appendix}
\newpage

\section{Data Creation}
\label{app:data_creating}

To construct our dataset, we sampled prompts from the Parti dataset and augmented them using GPT-4o. We randomly selected 50 samples from each of the following eight categories: Fine-grained Detail, Imagination, Simple Detail, Style and Format, Complex, Linguistic Structures, Perspective, and Quantity. For each category, we provided GPT-4o with the prompt: Create an alternative CSV with different prompts of similar style and complexity.

To ensure greater diversity, we repeated this process twice for Style and Format and Simple Detail, generating 100 examples for each. Finally, we manually reviewed all generated prompts to verify their diversity and coherence. In total, our dataset comprises 500 curated prompts.

\section{Attention Between Image and Text in Different Architectures}

To condition the generation process on a textual prompt, T2I models typically employ an attention mechanism.
There are two popular methods for achieving this: through cross-attention mechanism, used in models like Stable Diffusion 2 and Stable Diffusion XL, and Multimodal Diffusion Transformer (\textit{MM-DiT})~\citep{esser2024scaling} blocks, found in models such as FLUX and Stable Diffusion 3.
In the cross-attention mechanism, image patches are projected into query vectors $Q$ while text tokens are projected into key and value vectors $K$ and $V$.
Essentially, each image patch draws information from the text tokens based on the attention map $A$:
\begin{equation}
    A = \textit{softmax}(Q K ^\top / \sqrt{d_k}) ,
    \label{eq:cross_attention}
\end{equation}
where $d_k$ represents the dimensionality of the key vectors.
It is important to note that only the image patches extract information from the text tokens, while the text tokens remain constant throughout the computation process.
Alternatively, the \textit{MM-DiT} blocks implement a self-attention mechanism where both the image patches and text tokens are concatenated into a single set and then projected into $Q$, $K$ and $V$ vectors.
In this formulation, both the image and text draw information from each other, using the following attention map:
\begin{equation}
    A = \textit{softmax}([Q_{txt}, Q_{img}] [K_{txt}, K_{img}]^\top / \sqrt{d_k}) ,
    \label{eq:FLUX_attention}
\end{equation}
where $Q_{txt}$, $K_{txt}$ are the text query and key vectors, and $Q_{img}$, $K_{img}$ are the image query and key vectors.
Here, both the image patches and text tokens are updated after the operation.

\section{Models}
\label{app:models}
The models with frozen text encoders are:
\begin{enumerate}
    \item \textit{Stable Diffusion 2} employs a single frozen CLIP-based text encoder.
    \item \textit{Stable Diffusion 3} utilizes a combination of two frozen CLIP text encoders along with a frozen T5 encoder.
    \item \textit{FLUX} utilizes a frozen T5 text encoder and CLIP encoder. A key distinction between FLUX and Stable Diffusion models is that the latter incorporates a transformer architecture with self-attention to both the image and text latent representations in the diffusion process. This allows the diffusion model to modify text representations dynamically during the diffusion. We use the distilled version of FLUX - FLUX-Schnell.
\end{enumerate}

The models with trained text encoders:
\begin{enumerate}
    \item \textit{LDM} uses a BERT text encoder, which is trained jointly with the diffusion model on the image generation task.
    \item \textit{Lavi-Bridge} employs a LLaMA that is trained jointly with the diffusion model on the image generation task.
\end{enumerate}

\section{Technical Details}
All experiments were conducted using NVIDIA A100 GPUs with 8 cores, ensuring high computational performance and efficiency for our model evaluations. The total computational time across all experiments amounted to approximately 200 GPU hours.

\section{Additional Results}

\begin{table}[h]
    \centering
    \begin{tabular}{lcc}
        \toprule
        Model & Full & EOS \\
        \midrule
        SDXL & $0.34 \pm 0.036$ & $0.30 \pm 0.041$ \\
        SD2  & $0.33 \pm 0.033$ & $0.31 \pm 0.032$ \\
        \bottomrule
    \end{tabular}
    \caption{Average CLIP Scores for full representation vs images generated using \encoderMethod~masking all but the EOS token in Stable Diffusion XL and Strable Diffusion 2 models, both using CLIP as the text encoder.}
    \label{app:tab:eos_clip}
\end{table}

\paragraph{EOS Token Retains Most Semantic Information.}
\label{app:eos_clip}

When the CLIP text encoder is used, we observe a phenomenon where semantic information is concentrated in a special token—the EOS token. During CLIP’s training, this token serves as the primary objective for the text encoder, making its representation align with that of the image CLS token. As a result, semantic information naturally aggregates into this token. In Table \ref{app:tab:eos_clip}, we observe that the CLIP score of the EOS token is very high, nearly matching that of the image generated from the full representation. This observation has been noted in previous works, such as \citet{ding2024clip}.

\section{Complementary Results} \label{app:compl}
We have removed the error bars from Figure \ref{fig:main_results} for clarity. The standard deviations are listed in Table \ref{app:tab:main_std}.

\begin{table}[h!]
\centering
\begin{tabular}{lcccc}
\hline
Model & Clean Pads & Full & Pads & Prompt \\
\hline
FLUX & $0.039$ & $0.037$ & $0.036$ & $0.036$ \\
LDM & $0.033$ & $0.037$ & $0.043$ & $0.042$ \\
LLaMA-U & $0.034$ & $0.035$ & $0.034$ & $0.041$ \\
SD 2 & $0.037$ & $0.033$ & $0.037$ & $0.034$ \\
SD 3 & $0.039$ & $0.035$ & $0.046$ & $0.036$ \\
SDXL & $0.023$ & $0.036$ & $0.043$ & $0.039$ \\
\hline
\end{tabular}
\caption{Calculated standard deviation of CLIP scores for each model and different text encoder interventions. LLaMA-U stands for LLama-UNet. SD stands for Stable Diffusion.}
\label{app:tab:main_std}
\end{table}

\section{Qualitative Examples} \label{app:qual}
The following figures provide visual examples illustrating the impact of padding tokens in the T2I pipeline, highlighting some key findings from our analysis.

\begin{figure*}[h!]  %
  \centering
  \includegraphics[width=.8\textwidth]{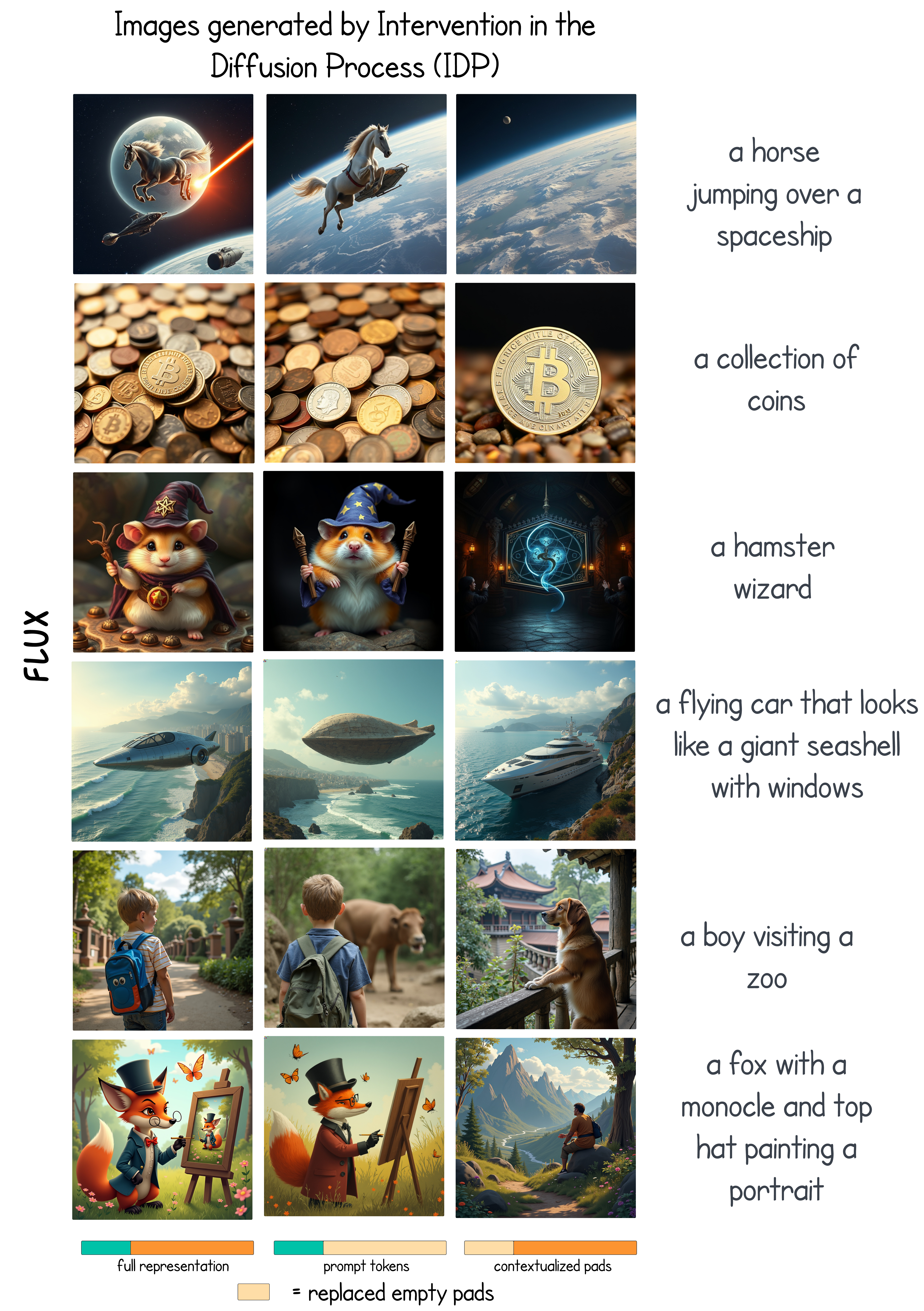}  %
  \caption{Additional examples of images generated from different segments of the input prompt using~\diffusionMethod. Description of each column, from left to right: (1) An image generated using the full prompt (both prompt tokens and padding tokens encoded together), (2) An image generated using only the prompt tokens and \textcolor{blank_pads}{clean padding tokens} that were not encoded with the prompt, (3) An image generated using only the \textcolor{prompt_pads}{padding tokens} encoded with the prompt, while the prompt tokens were replaced with \textcolor{blank_pads}{clean pad tokens}. See Figure~\ref{fig:diffusion_causal} for further technical details.}
  \label{fig:causal_diffusion_examples}
\end{figure*}

\begin{figure*}[h!]  %
  \centering
  \includegraphics[width=.82\textwidth]{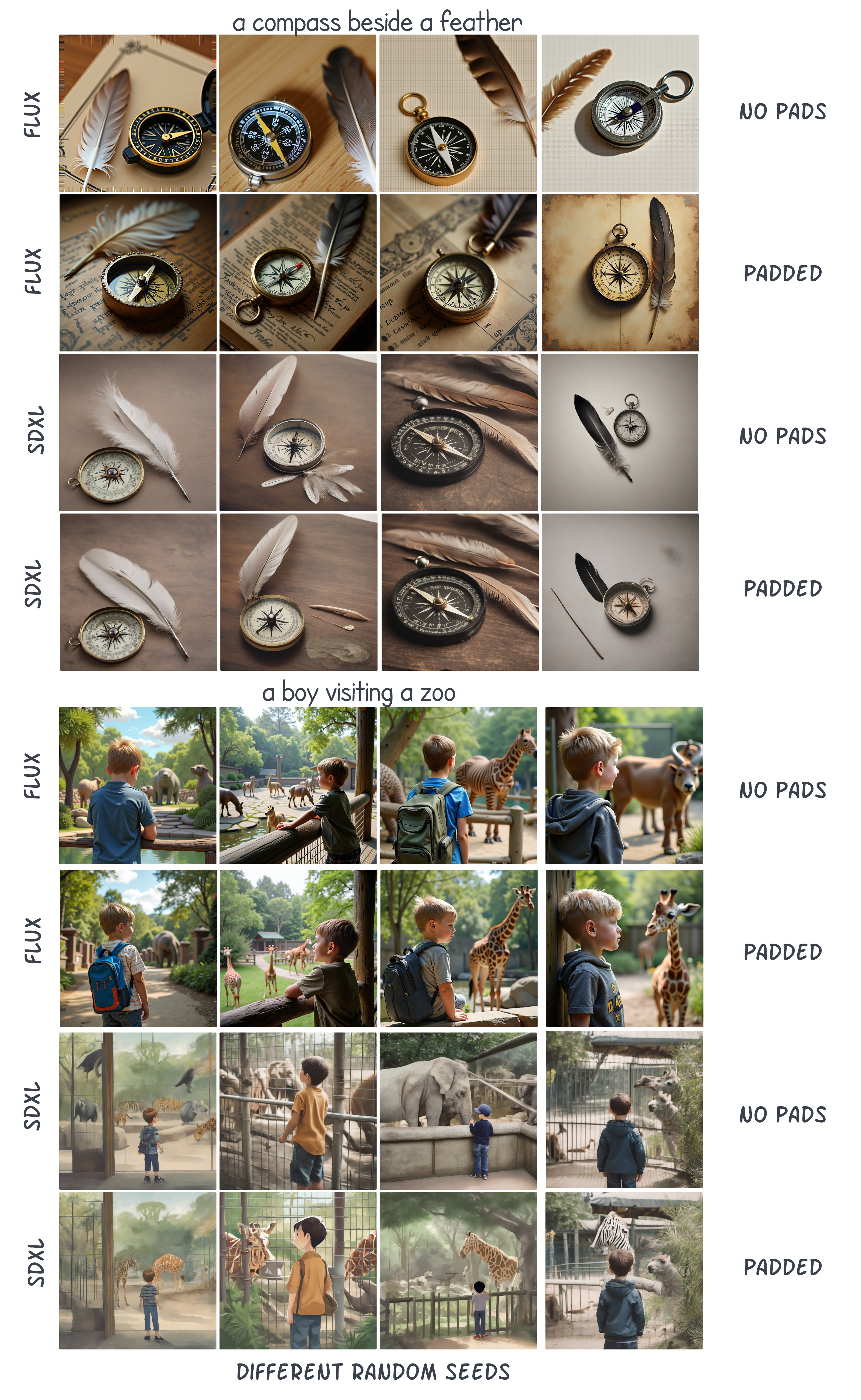}  %
  \caption{Examples of images generated from the same prompts with maximum padding* and without padding in Stable Diffusion XL and FLUX. Images generated by Stable Diffusion XL maintain consistent quality, while produced by FLUX without padding often miss key details. For example, given the prompt \textit{``a compass beside a feather,''} images with padding typically include textured paper with text or a manuscript. In contrast, for the prompt \textit{``a boy visiting a zoo,''} images generated without padding result in vague animal shapes (first column) or hybrids, such as a mix between a giraffe and a horse (third image). However, adding padding leads to more visually coherent animals. *Maximum padding length is defined as the default number of padding tokens for each model: 77 for SDXL and 512 for FLUX.
 }
  \label{fig:max_len}
\end{figure*}

\newpage

\end{document}